\documentclass[conference]{IEEEtran}
\IEEEoverridecommandlockouts
\usepackage[T1]{fontenc} 
\usepackage{cite}
\usepackage{amsmath,amssymb,amsfonts}
\usepackage{algorithmic}
\usepackage{graphicx}
\usepackage{booktabs} 
\usepackage{subcaption}
\usepackage{siunitx}
\usepackage{textcomp}
\usepackage{xcolor}
\usepackage{url}
\usepackage{placeins}
\def\BibTeX{{\rm B\kern-.05em{\sc i\kern-.025em b}\kern-.08em
    T\kern-.1667em\lower.7ex\hbox{E}\kern-.125emX}}

\urldef{\projecturl}\url{https://nemantor.github.io/sparse-3d-traversal-website/}

\newif\ifanonymous
\anonymousfalse

\begin{document}
\bstctlcite{BSTcontrol}

\title{Learning Agile Perceptive Traversal of\\Sparse 3D Structures for Humanoids}

\ifanonymous
  \author{\IEEEauthorblockN{Anonymous ICRA 2027 submission}
  \IEEEauthorblockA{Paper ID \textbf{XXXX}}}
\else
  \author{%
  \IEEEauthorblockN{Efe Ongan$^{1}$, Chong Zhang$^{1,2}$, Boyang Sun$^{3}$, Andrei Cramariuc$^{1}$, Cesar Cadena$^{1}$, Marco Hutter$^{1}$}
  \thanks{$^{1}$Robotic Systems Lab, $^{2}$ETH AI Center, $^{3}$Computer Vision and Geometry Group, ETH Zurich, Switzerland.}
  \thanks{Project website: \projecturl}%
}
\fi

\maketitle
\thispagestyle{empty}
\pagestyle{empty}

\begin{abstract}
Traversing sparse 3D structures requires humanoid robots to perceive thin,
overhanging geometry while executing agile, accurate whole-body
motions. We study this problem through monkey-bar traversal, where the robot
must jump to the structure, traverse it through sparse bar interactions, and
land safely. For this task, we present a reinforcement-learning-based perceptive control
system that operates directly on observations from a head-mounted
solid-state lidar. To extract task-relevant geometry from the sparse
returns, the policy consumes the raw lidar scan through an attention-based
encoder with recurrent memory. This policy is obtained by a phase-scheduled
teacher--student pipeline that combines privileged experts for jumping up,
brachiating, and jumping down. For transfer to hardware, we model lidar
noise, battery-voltage sag, and actuator thermal limits, and equip the
humanoid with passive hook end-effectors for robust bar interaction. On
hardware, the resulting policy completes the full
jump-up$\to$brachiation$\to$jump-down sequence in 14 of 15 trials across
three bar configurations and reaches brachiation speeds up to 0.5~m/s.
Beyond brachiation, the same perception backbone supports a separately
trained policy that ducks beneath thin overhead obstacles with 2~cm
cross-sections.
\end{abstract}
\vspace{-5mm}

\section{Introduction}
Legged robots often rely on intermediate spatial representations for
whole-body control and locomotion, such as 2.5D elevation
maps~\cite{miki2022elevationmappinglocomotionnavigation, zhang2026ame2agilegeneralizedlegged}
and voxel grids~\cite{ben2025gallantvoxelgridbasedhumanoid}, which aid
generalization and learning efficiency but at a cost: elevation maps lose
thin and overhanging structures, and voxel grids pay steeply in memory and
compute as resolution grows. An alternative line of work consumes raw
sensor data---depth images or point
clouds---directly~\cite{yang2025spatiallyenhancedrecurrentmemorylongrange, rudin2025parkourwildlearninggeneral, wang2025omni},
but has so far addressed dense terrain and large obstacles; agile
interaction with sparse, thin structures remains largely unexplored.

We close this gap through brachiation on monkey bars, an extreme instance
of this setting that stresses three axes at once: fine-grained perception
of the thin bars, hard exploration of precise contact sequences, and
agile whole-body control under tight hardware limits.

\begin{figure}[!t]
  \centering
  \includegraphics[width=0.8\columnwidth]{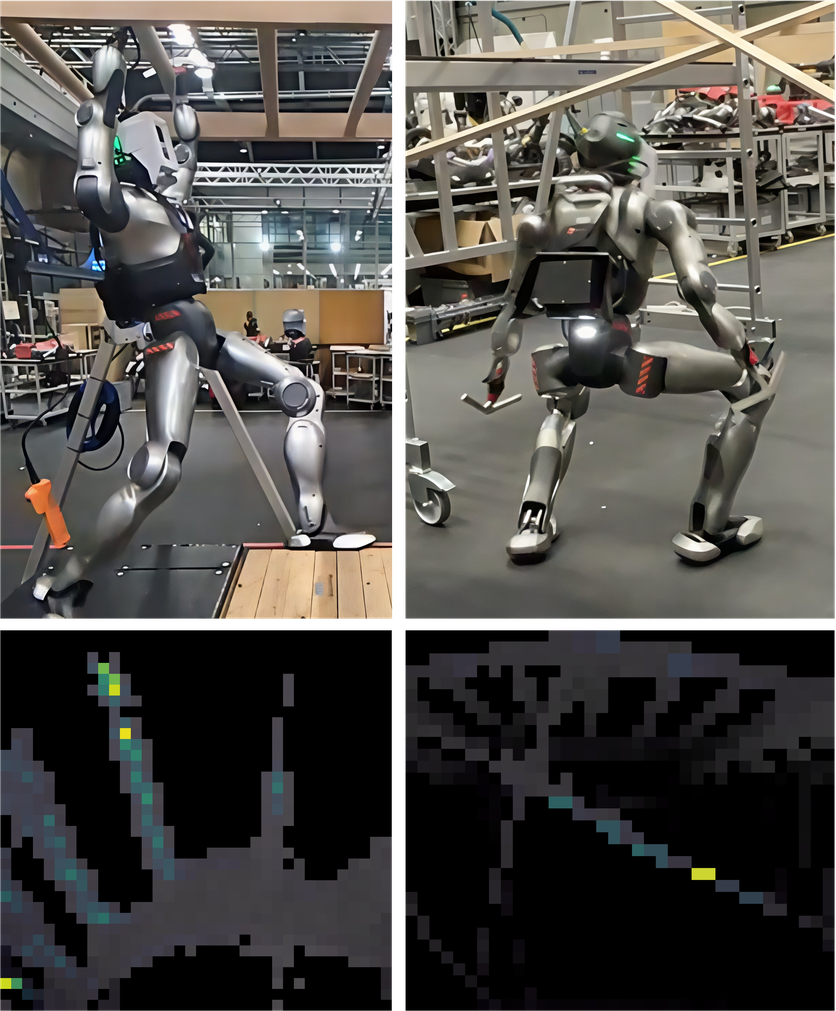}
  \caption{Hardware deployment: jumping onto the bar (left) and ducking
    beneath thin overhead bars (right); bottom: the policies' learned
    attention pattern for lidar observations, where task-related returns are focused on.}
    \vspace{-4mm} 
  \label{fig:hero}
\end{figure}

\begin{figure*}[!t]
  \centering
  \includegraphics[width=\textwidth]{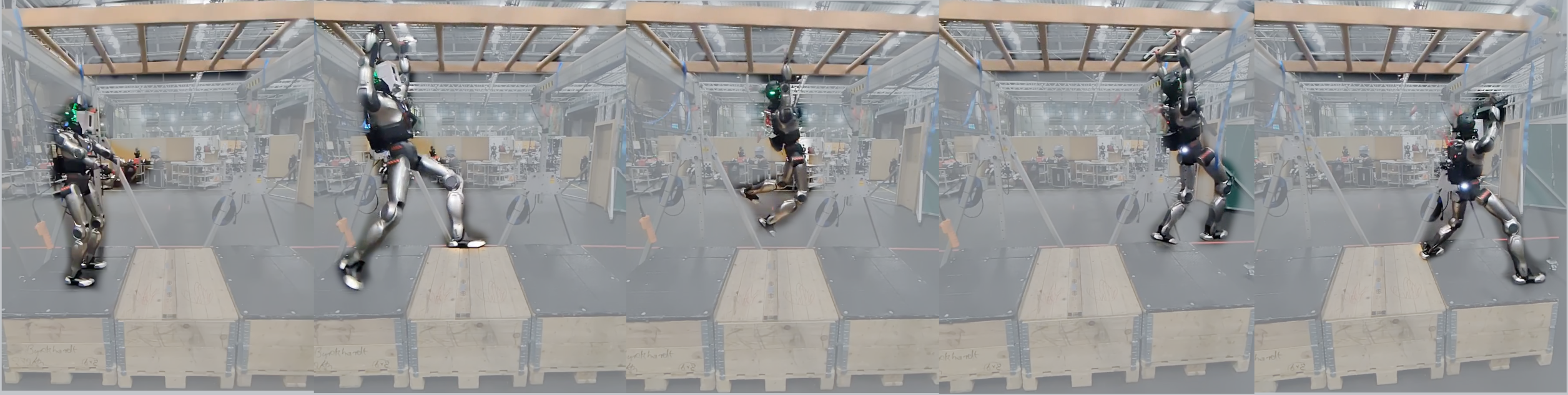}\\[6pt]
  \includegraphics[width=\textwidth]{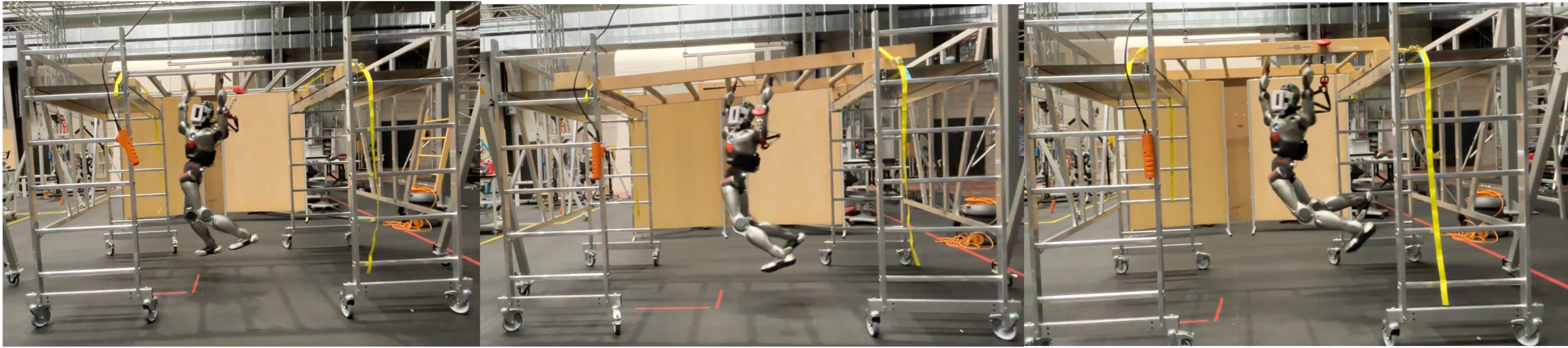}
  \caption{Our learned perceptive controller traverses different monkey bar setups.}
  \vspace{-3mm}
  \label{fig:motion}
\end{figure*}

Our system addresses each axis in turn. For perception, since the bars yield only
a few, intermittent returns in each raw scan of a head-mounted
solid-state lidar, we use an attention-based encoder, shown effective on elevation-maps~\cite{zhang2026ame2agilegeneralizedlegged, he2025attentionbasedmapencodinglearning},
to select these task-relevant returns, and a recurrent memory to integrate
them over time. To facilitate exploration, we train privileged subtask experts and distill them,
under a phase schedule, into a single perceptive student. For hardware performance, we equip a PM-01 humanoid~\cite{engineaiPM01_ENGINEAI} with passive hook end-effectors, and model battery-voltage sag, actuator thermal limits, and lidar noise for sim-to-real transfer
(Fig.~\ref{fig:hero}). To our knowledge, this is the first demonstration
of a humanoid perceiving and executing the complete
jump-up--brachiation--jump-down sequence using onboard sensing
(Fig.~\ref{fig:motion}).

We summarize our contributions as follows:
\begin{itemize}
    \item An end-to-end attention-based perceptive control system in which a humanoid traverses
      sparse 3D structures directly with solid-state lidar observations;
    \item A phase-scheduled multi-teacher distillation scheme with PPO post-refinement, which combines subtask experts into a single deployable policy;
    \item Models of battery voltage sag, actuator thermal limits, and lidar
      noise that enable sim-to-real transfer of explosive whole-body maneuvers;
    \item Hardware validation across diverse bar configurations, together with a demonstration that the same perception backbone supports ducking under thin overhead obstacles without architectural modifications.
\end{itemize}

\section{Related Work}
\textbf{Perceptive locomotion.}
Learned controllers widely consume intermediate representations:
robot-centric elevation maps~\cite{miki2022elevationmappinglocomotionnavigation, miki2022learning, zhang2026ame2agilegeneralizedlegged, he2025attentionbasedmapencodinglearning}
discard overhangs and thin structures, while voxel
grids~\cite{ben2025gallantvoxelgridbasedhumanoid} recover full 3D geometry
at a resolution-dependent memory and compute cost. Map-free approaches feed raw
exteroception directly to the policy: egocentric depth for quadruped and
humanoid
parkour~\cite{agarwal2022leggedlocomotionchallengingterrains, cheng2023parkour, zhuang2023robotparkour, zhuang2024humanoidparkourlearning},
recurrent depth
encoders~\cite{yang2025spatiallyenhancedrecurrentmemorylongrange}, and raw
lidar point clouds for collision-avoidance locomotion on
humanoids~\cite{wang2025omni, wang2025safecomfortable}. 
Similarly, our work trains policies that directly consume raw lidar observations, while targeting contact-accurate interaction with centimeter-scale sparse structures that previous works did not cover.

\textbf{Privileged distillation.}
Teacher--student training~\cite{chen2019learningcheating} decouples exploration
from partial observability; it is widely used for blind
locomotion~\cite{Lee_2020, kumar2021rmarapidmotoradaptation} and perceptive
locomotion~\cite{miki2022learning}, typically via
DAgger~\cite{ross2011reductionimitationlearningstructured}. Closest to ours
are methods that distill multiple specialists into one policy via DAgger,
with~\cite{rudin2025parkourwildlearninggeneral, zhao2026ladderman} or
without~\cite{zhuang2023robotparkour} subsequent RL fine-tuning;
ANYmal Parkour~\cite{hoeller2023anymalparkourlearningagile} instead keeps
its skills as separate policies and switches between them. We differ in
scheduling the active teacher by task phase within a single episode and in
normalizing advantages per phase, in the spirit of
PopArt~\cite{hessel2019popart}, to handle heterogeneous reward scales.

\textbf{Sim-to-real modeling.}
Accurate actuator models are central to transferring dynamic skills,
learned from data~\cite{hwangbo2019learning}, targeted at extreme humanoid
motion~\cite{wang2026omnixtreme}, or physics-inspired with parameters fitted
on the robot~\cite{mujoco_dcmotor, mueller2026olaf}. Beyond actuation,
realistic lidar simulation has modeled sensor-specific nonidealities such as
ray drop and return intensity from real-world data for automotive spinning
lidars~\cite{manivasagam2020lidarsimrealisticlidarsimulation, guillard2022learningsimulaterealisticlidars}
and for legged-robot lidars~\cite{wang2025omni}; depth-based locomotion
policies instead use heuristic edge
corruption~\cite{rudin2025parkourwildlearninggeneral}. We build on these
lines by adapting battery-voltage and actuator-thermal proxies to explosive
whole-body maneuvers and by identifying the dominant noise processes of a
specific solid-state lidar, the RoboSense E1R \cite{robosenseRoboSenseSafer}, for training and validation.

\textbf{Dynamic traversal and brachiation.}
Ladder climbing has been addressed with model-based multi-contact planning on humanoids~\cite{vaillant2016multicontact} and learned controllers on quadrupeds~\cite{vogel2025robustladderclimbingquadrupedal}. More recently, LadderMan~\cite{zhao2026ladderman} demonstrated perceptive humanoid ladder climbing with a depth-based visuomotor policy generalizing across ladder geometries. Brachiation has been studied on specialized two- and three-link
platforms~\cite{saito1994, Javadi_2023, grama2024ricmonk}, the more recent of
which use passive hook grippers,
in simulation through simplified-model
imitation~\cite{10.1145/3528233.3530728}, and recently with zero-shot
sim-to-real transfer on a life-sized dual-arm robot using waypoint-guided
RL~\cite{iwata2026robust}. These systems either are purpose-built
brachiators or assume known bar positions; none perceives the bars onboard
or combines brachiation with jumping up to and down from the structure on a
general-purpose humanoid.

\section{Hardware}
\subsection{Passive Hook End-Effectors}
The end-effector must fulfill four roles: carrying the robot's full weight
under the dynamic loads of swinging and catching; tolerating the contact-placement
error of a learned controller; releasing from a bar without demanding large arm
torques; and surviving high-force impacts.

\begin{figure}[!t]
  \centering
  \begin{subfigure}{\columnwidth}
    \centering
    \includegraphics[width=\columnwidth]{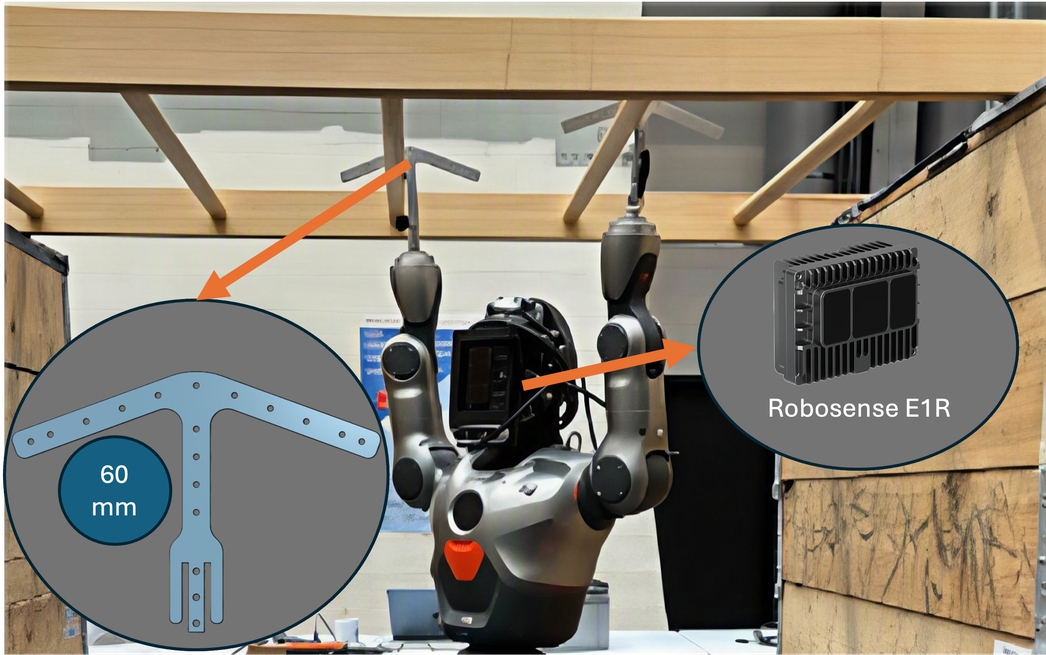}
    \caption{}
    \label{fig:style_hook}
  \end{subfigure}\\[3pt]
  \begin{subfigure}{\columnwidth}
    \centering
    \includegraphics[width=\columnwidth]{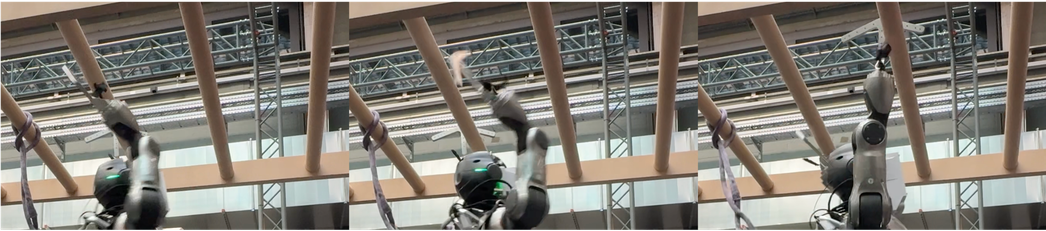}
    \caption{}
    \label{fig:style_yaw}
  \end{subfigure}\\[3pt]
  \begin{subfigure}{\columnwidth}
    \centering
    \includegraphics[width=\columnwidth]{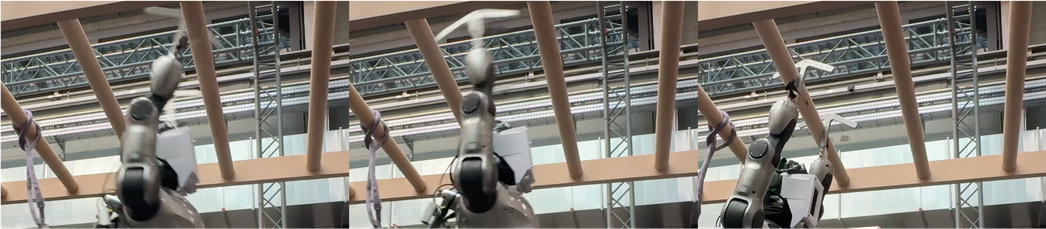}
    \caption{}
    \label{fig:style_backward}
  \end{subfigure}
  \caption{(a) Close-up of the modified PM-01 with the passive hook
    and RoboSense E1R lidar highlighted. (b) Use of the hand yaw joint for disengaging from the
    bars. (c) Backward movement enabled by the symmetric hook
    construction.}
  \label{fig:brachiation_style}
\end{figure}

Our hook (Fig.~\ref{fig:style_hook}) is a single water-jet-cut
stainless-steel plate that replaces the hand entirely. The opening admits a
\SI{60}{\milli\metre} circle, wide relative to the $1$--$3$\,cm bar radii
of Sec.~\ref{sec:terrain}, giving the policy tolerance to contact-placement
error in the swing plane.

Disengagement is achieved using the wrist-yaw joint
(Fig.~\ref{fig:style_yaw}): rotating the hook out of the bar plane
releases the contact using wrist-yaw torque alone, without lifting the body.
This reduces the actuator effort required during release and helps keep the
actuators within their thermal limits. The symmetric geometry further enables
bidirectional traversal and multiple contact modes
(Fig.~\ref{fig:style_backward}). Its simple shape can also be represented
accurately using primitive box colliders, reducing collision-detection cost in
simulation.

\subsection{Head Mounted Solid-State Lidar}
The sensor must resolve bars of $1$--$3$\,cm radius at jumping distance,
maintain accuracy during high accelerations, and keep both the next bar and
the ground in view without actuated gaze control. We use the RoboSense E1R~\cite{robosenseRoboSenseSafer}, a
solid-state lidar with an integrated IMU, and a dense
$192 \times 144$ scan pattern over a $120^{\circ} \times 90^{\circ}$ field of
view. We specifically favor a
solid-state sensor because its electronic scan pattern reduces the
motion-induced scanning distortion that can arise in mechanically scanning lidars during fast whole-body motions. A sample return is shown in
Fig.~\ref{fig:lidar_real}. The policy consumes a decimated subset of this pattern directly.

\begin{figure}[!t]
  \centering
  \includegraphics[trim=0 7cm 0 0, clip, width=\columnwidth]{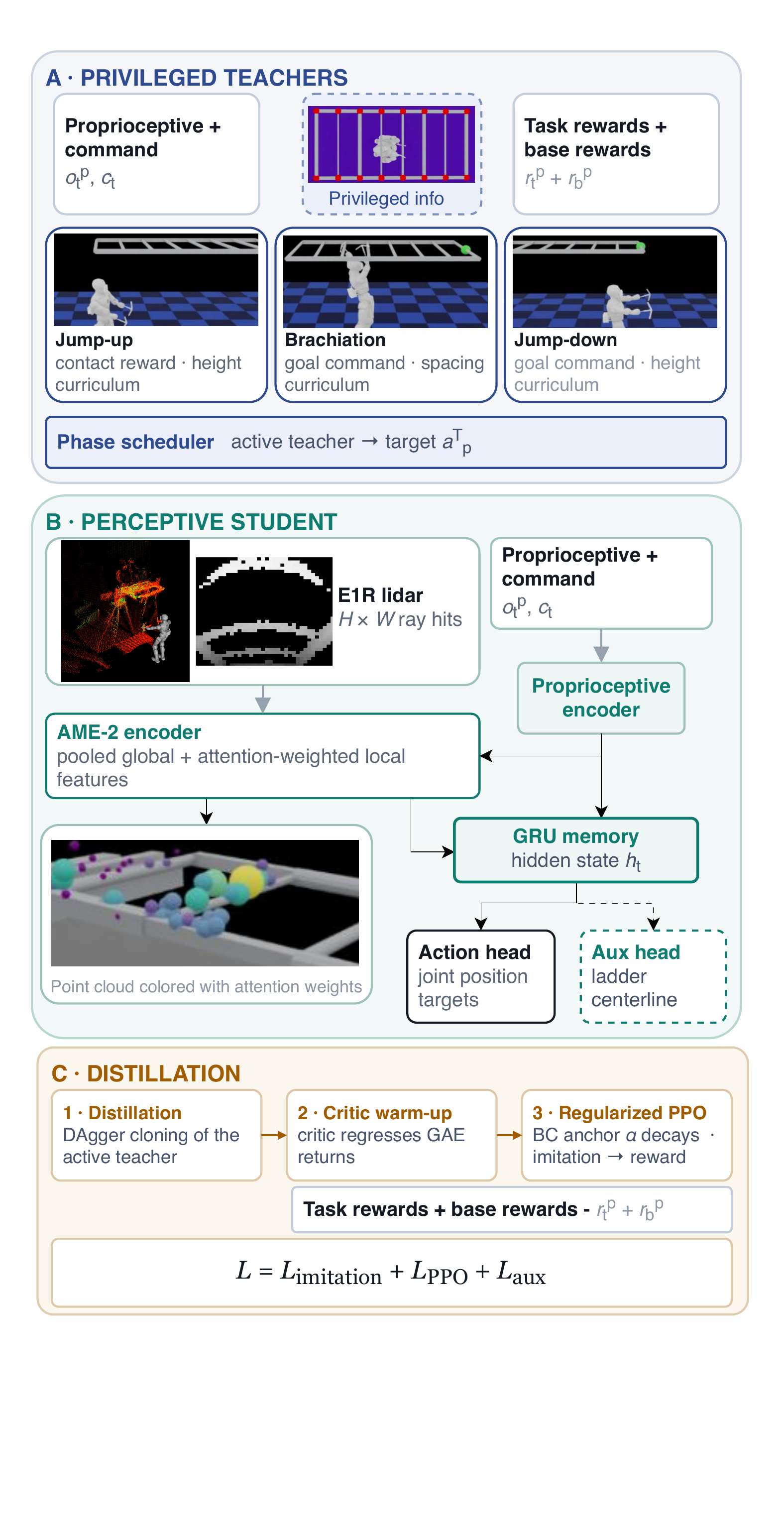}
  \caption{System overview. (A) Privileged subtask teachers and the phase
    scheduler. (B) Perceptive student: grid-structured attention encoder,
    GRU memory, action and auxiliary heads. (C) Three-stage multi-teacher
    distillation.}
   \vspace{-5mm}
  \label{fig:full_architecture}
\end{figure}

\section{Learning Approach}

\subsection{Overview}
The exploration problem,
discovering contact modes for jumping up, brachiating, and jumping down, is
solved by reinforcement learning of privileged experts with access to ground-truth bar positions. The
perception problem is solved by a single student that observes only onboard
sensing and learns to reproduce the expert behaviors from raw lidar, first
through phase-scheduled distillation and then through PPO refinement
(Fig.~\ref{fig:full_architecture}). Only the exploration stage is
specific to brachiation; Sec.~\ref{sec:duck_policy} reuses the perception
components unchanged for a second task of avoiding thin sparse obstacles.

Policies are trained in IsaacLab~\cite{mittal2025isaaclab} with the
RSL-RL~\cite{schwarke2025rslrl} implementation of
PPO~\cite{schulman2017proximalpolicyoptimizationalgorithms}.

\subsection{Privileged Teachers}
\label{sec:teachers}
We train three privileged teachers, each specializing in one subtask:
brachiation, jumping-up, and jumping-down. All teachers receive a privileged
bar observation consisting of the endpoints of the bars on the terrain
(Sec.~\ref{sec:terrain}). The brachiation teacher is conditioned on
goal-position commands, while the jumping-up and jumping-down teachers are
conditioned on contact-based rewards. This encourages the policies to
discover distinct contact modes without human motion priors. All teachers are MLP policies. Beyond the bar endpoints, teachers and the
critic receive privileged state unavailable on hardware: contact states,
base velocity, and the battery and thermal states of
Sec.~\ref{sec:battery_thermal}.
For the obstacle avoidance task, the privileged teacher follows the training formulation of~\cite{zhang2026ame2agilegeneralizedlegged} with an additional upward-directed raycast channel on its height map to register overhead obstacles; this height map is teacher-only and never observed by the deployed student. 

\subsection{Training Terrain and Curriculum}
\label{sec:terrain}
Policies are trained on a procedurally generated ladder whose geometry is
randomized per environment. For robustness
against spatial disturbances, random objects are additionally scattered around
the ladder. For training the jump-up and
jump-down policies, a curriculum of increasing ladder height is used; for
brachiation, a curriculum with increased bar spacing and reduced bar width. The ranges used during curriculum generation are in Table~\ref{tab:domain_rand}.

\subsection{Reward Design}
\label{sec:rewards}
The reward consists of two groups: task terms
specific to each teacher, and limit terms that protect the hardware and enforce physically plausible motion (Table~\ref{tab:rewards}). The limit reward group utilizes the
platform limits of Sec.~\ref{sec:battery_thermal}: the thermal integrator
and the battery voltage are penalized directly, and the summed torque per
limb is budgeted to protect the cable harness. Joint targets additionally
pass through a filter that clips them to a soft limit band; the clipped
magnitude is penalized so the policy learns to respect the limits rather
than lean on the filter. The position-conditioned brachiation and 
obstacle avoidance tasks use the \cite{zhang2026ame2agilegeneralizedlegged}
task-reward formulation with additional dense hand progress rewards for 
aiding exploration. The jumping-up and jumping-down tasks use the same formulation of position-conditioned tasks but replace target goals with contact position targets. 

  \begin{table}[!t]
  \centering
  \caption{Reward terms. The base block is shared by all tasks; each
    teacher adds its task block.}
  \label{tab:rewards}
  \footnotesize
  \setlength{\tabcolsep}{2pt}
  \begin{tabular}{llr}
    \toprule
    Term & Definition & Weight \\
    \midrule
    \multicolumn{3}{l}{\textit{Base (shared)}} \\
    Undesired contact & body contacts $>$ \SI{1}{\newton} & $-2$ \\
    Contact sliding & hand and foot slip while loaded & $-0.1$ \\
    Filter clipping & clipped joint targets ($5\times$ for ankles) & $-0.1$ \\
    Joint pos.\ limit & excursion beyond limit & $-100$ \\
    Joint vel.\ limit & excursion above $0.9\times$ limit & $-1$ \\
    Joint torque limit & excursion beyond limit & $-10^{-3}$ \\
    Thermal & arm integrator above $0.8$ & $-0.1$ \\
    Battery & voltage hinge below \SI{40}{\volt} & $-0.1$ \\
    Limb torque budget & hinge on limb $\sum_j |\tau_j|$ & $-10^{-4}$ \\
    \midrule
    \multicolumn{3}{l}{\textit{Task: brachiation}} \\
    Goal tracking & $1/(1+d_{xy}^{2}/0.3)$ & $10$ \\
    Progress & $+1$ per \SI{2}{\centi\metre} closest approach & \\
    & \quad (base and each hand) & $3\times5$ \\
    Stagnation & termination on stagnation & $-200$ \\
    \midrule
    \multicolumn{3}{l}{\textit{Task: jump-up}} \\
    Hand / torso tracking & exp-kernel at nearest bar pos error & $5$; $5$ \\
    Progress & $+1$ per \SI{2}{\centi\metre} (hands, torso) & $3\times2$ \\
    Contact at bar & hand contact within target radius & $2$ \\
    Grasp load & downward hand force at bar & $1$ \\
    Hold / hang bonus & sustained grasp; stable hang & $1$; $5$ \\
    \midrule
    \multicolumn{3}{l}{\textit{Task: jump-down}} \\
    Landing contact & feet contact below bar height & $5$ \\
    Root height & exp-kernel at landing height pos error & $3$ \\
    Descent braking & foot speed near ground & $-0.25$ \\
    Bar contact after release & contacts $>$ \SI{1}{\newton} & $-1$ \\
    Heading & misalignment to bar row & $-2$ \\
    Stable landing & upright at episode end & $100$ \\
    Out of bounds & termination & $-200$ \\
    \midrule
    \multicolumn{3}{l}{\textit{Task: ducking (formulation of~\cite{zhang2026ame2agilegeneralizedlegged})}} \\
    \bottomrule
  \end{tabular}
  \vspace{-5mm}
  \end{table}

\begin{figure*}[!t]
  \centering
  \begin{subfigure}[t]{0.49\textwidth}
    \centering
    \includegraphics[width=\linewidth]{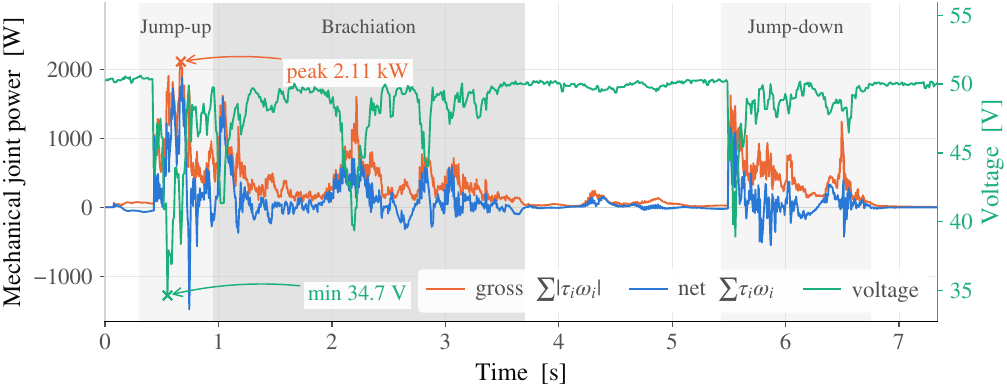}
    \caption{Estimated mechanical joint power.}
    \label{fig:power}
  \end{subfigure}\hfill
  \begin{subfigure}[t]{0.49\textwidth}
    \centering
    \includegraphics[width=\linewidth]{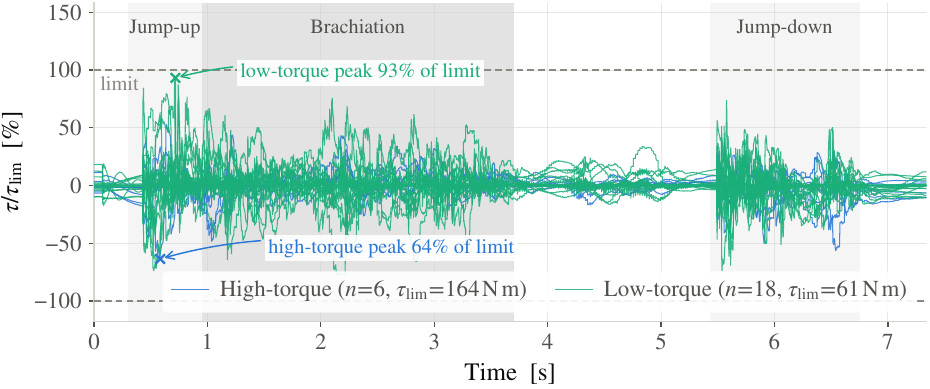}
    \caption{Limit-normalized estimated actuator torques.}
    \label{fig:torque}
  \end{subfigure}
  \caption{Hardware measurements over a full jump-up/brachiation/jump-down
    sequence (phases shaded). Summed mechanical joint power (a) peaks at
    \SI{2.11}{\kilo\watt} during the jump-up with a battery voltage of \SI{34.7}{\volt} at minimum; the 18 low-torque
    actuators including 10 arm joints (b) approach their limits during voltage sags,
    motivating the models of Sec.~\ref{sec:battery_thermal}.}
  \label{fig:power_torque}
\end{figure*}

\subsection{Multi-Teacher Distillation}
\label{sec:distill}
The teachers are distilled into a single perceptive student in three stages,
with the active teacher $p$ scheduled according to task progression so that the
corresponding teacher action $a^{T}_{p}$ and reward are forwarded to the
student.

\textbf{(i) Distillation}: pure behavior cloning
(DAgger~\cite{ross2011reductionimitationlearningstructured}) with an MSE loss through truncated
BPTT.

\textbf{(ii) Critic warmup}: prevents instabilities at the transition to reinforcement learning.
The critic regresses unclipped GAE returns, while actor BC updates continue as in \cite{rudin2025parkourwildlearninggeneral}.

\textbf{(iii) Regularized PPO}: we gradually transition from imitation to the
task objective using the clipped PPO surrogate and a decaying BC anchor,
allowing the student to adapt to its perceptual limitations. Since reward
scales differ across phases, we standardize advantages per phase and weight
the value loss by clipped inverse return variance~\cite{hessel2019popart},
preventing any phase from dominating the updates.

\subsection{Policy Architecture}
\label{sec:policy_arch}
The student observes proprioception $o^\mathrm{p}_t$ (positions and
velocities of all actuated joints except the head-yaw joint, IMU angular velocity and projected
gravity, and the previous action, with a $4$-frame history), a command $c_t$
(the planar goal position and a jump-down trigger), and the E1R returns
$o^\mathrm{e}_t \in \mathbb{R}^{36\times35\times4}$ ($36\times48\times4$ for ducking), refreshed at
\SI{10}{\hertz}, where each node of the native scan grid holds the hit
position relative to the sensor and its range. Brachiation crops rays to the upper-half workspace, while ducking retains the full field of view. To encode $o^\mathrm{e}_t$ we
adapt the attention-based map encoder of
AME-2~\cite{zhang2026ame2agilegeneralizedlegged} to the point cloud directly:
instead of processing the returns as an unordered
set~\cite{10.5555/3295222.3295263}, the cloud is kept on its 2D grid. A GRU\cite{cho2014learning}
memory module is fed the perception features together with an embedding of
$o^\mathrm{p}_t$ and $c_t$; its hidden state is decoded by an MLP into the
action $a_t \in \mathbb{R}^{23}$, joint position targets excluding head-yaw joint at \SI{50}{\hertz}
tracked by joint PD controllers.

\subsection{Auxiliary Guidance Losses}
\label{sec:centerline_probe}
To provide a strong gradient signal for the recurrent perceptive backbone during
distillation, we introduce an auxiliary bar-centerline prediction loss by learning a decoder on the GRU hidden state that predicts the closest bar's relative position and orientation.

\subsection{Ducking Under Thin Obstacles}
\label{sec:duck_policy}
We train a second, standalone ducking policy (Fig.~\ref{fig:ducking}) conditioned to follow goal positions while avoiding overhanging obstacles. It is distilled
from the height-map-based privileged teacher of Sec.~\ref{sec:teachers},
but the student itself reuses the AME-2 encoder and GRU memory of
Sec.~\ref{sec:policy_arch} unchanged, consumes only the raw E1R observation,
and is trained under the lidar noise model of Sec.~\ref{sec:lidar_model}.
\section{Sim-to-Real Transfer}
Sim-to-real transfer requires modeling the physical effects that dominate
at the platform's limits (Fig.~\ref{fig:power_torque}): battery voltage
sag, actuator thermal limits, and lidar noise. We randomize the parameters
according to Table~\ref{tab:domain_rand}.

\begin{table}[!t]
\centering
\caption{Domain randomization.}
\label{tab:domain_rand}
\footnotesize
\setlength{\tabcolsep}{3pt}
\begin{tabular}{llc}
  \toprule
  Parameter & Scope & Range \\
  \midrule
  \multicolumn{3}{l}{\textit{Terrain}} \\
  Bar radius [m] & ladder & $(0.01, 0.03)$ \\
  Bar width [m] & ladder & $(0.7, 1.2)$ \\
  Bar spacing [m] & ladder & $(0.25, 0.5)$ \\
  Bar height [m] & curriculum & $(1.6, 2.1)$ \\
  Tilt angle [$^\circ$] & ladder & $(-5, 5)$ \\
  \midrule
  \multicolumn{3}{l}{\textit{Physics (startup)}} \\
  Static / dynamic friction & all bodies & $(0.3, 2.0)$ \\
  Restitution & all bodies & $(0.0, 0.5)$ \\
  Link mass & all links, scale & $(0.8, 1.2)$ \\
  Torso mass [kg] & additive & $(-2, 5)$ \\
  CoM offset [m] & links / torso & $\pm0.01$ / $\pm0.05$ \\
  PD gains & scale & $(0.8, 1.2)$ \\
  Joint armature, friction & scale & $(0.1, 2.0)$ \\
  \midrule
  \multicolumn{3}{l}{\textit{Actuation (reset)}} \\
  Effort / velocity limit & scale & $(0.8, 1.0)$ \\
  Torque scale & all joints & $(0.95, 1.05)$ \\
  Torque bias [N\,m] & all joints & $\pm1$ \\
  Motor zero offset [rad] & all joints & $\pm0.035$ \\
  Battery sag $k_\mathrm{sag}$ & scale & $(0.5, 1.5)$ \\
  \midrule
  \multicolumn{3}{l}{\textit{Disturbances}} \\
  Torso force [N] / torque [N\,m] & reset, every 0--20\,s & $\pm25$ / $\pm10$ \\
  Velocity push, lin.\ [m/s] & every 0--5\,s & $\pm0.5$ ($z$: $\pm0.25$) \\
  Velocity push, ang.\ [rad/s] & every 0--5\,s & $\pm0.5$ \\
  Initial base pose [m] & reset & $\pm0.5$ \\
  Initial base velocity [m/s, rad/s] & reset & $\pm0.2$, $\pm0.5$ \\
  Initial joint offset [rad, rad/s] & reset & $\pm0.2$, $\pm0.5$ \\
  \bottomrule
\end{tabular}
\vspace{-5mm}
\end{table}

\begin{figure*}[!t]
  \centering
  \begin{subfigure}[t]{0.24\textwidth}
    \centering
    \includegraphics[width=\textwidth]{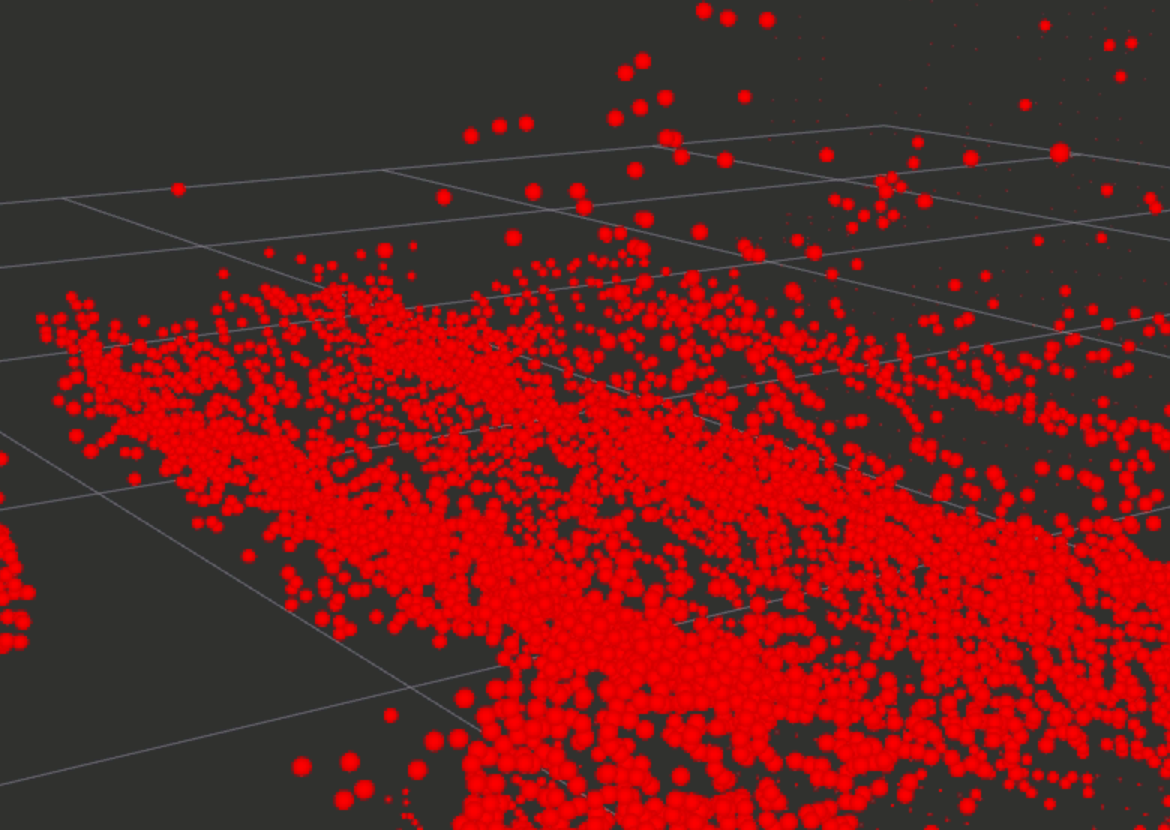}
    \caption{Hardware}
    \label{fig:lidar_real}
  \end{subfigure}\hfill
  \begin{subfigure}[t]{0.24\textwidth}
    \centering
    \includegraphics[width=\textwidth]{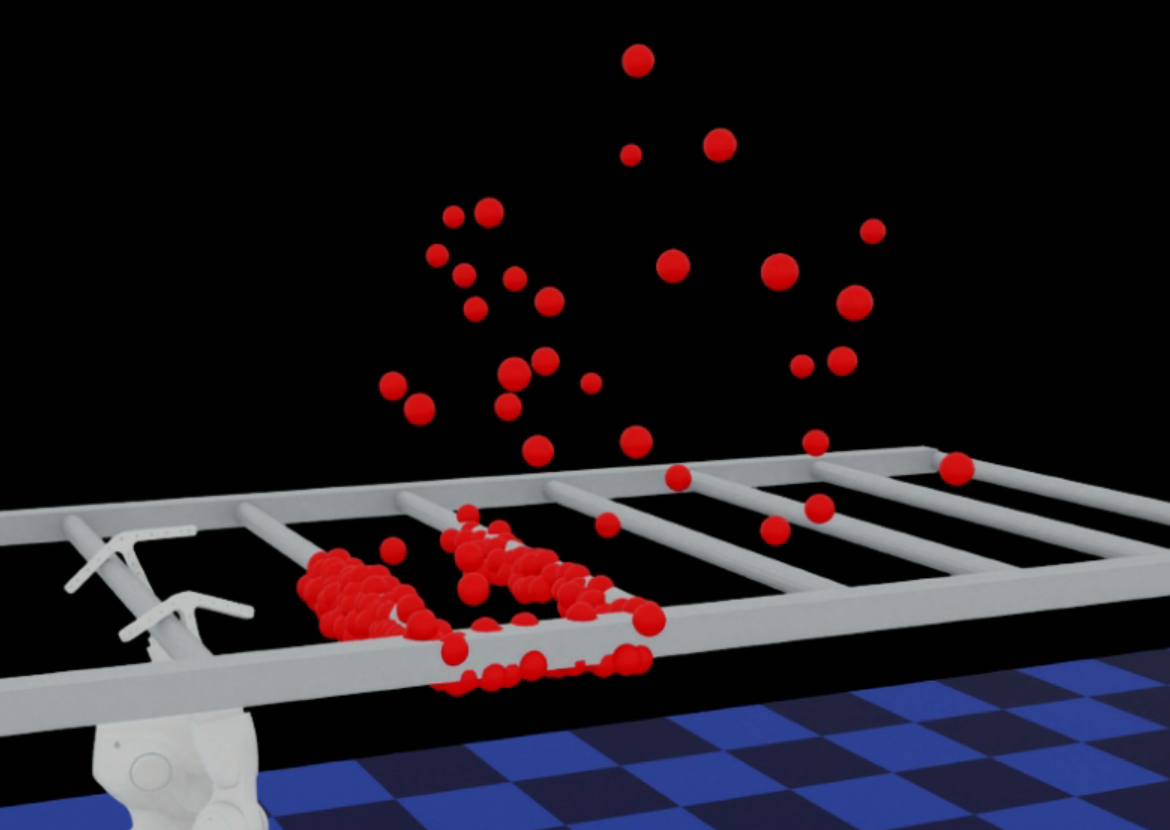}
    \caption{IsaacLab}
    \label{fig:lidar_isaaclab}
  \end{subfigure}\hfill
  \begin{subfigure}[t]{0.24\textwidth}
    \centering
    \includegraphics[width=\textwidth]{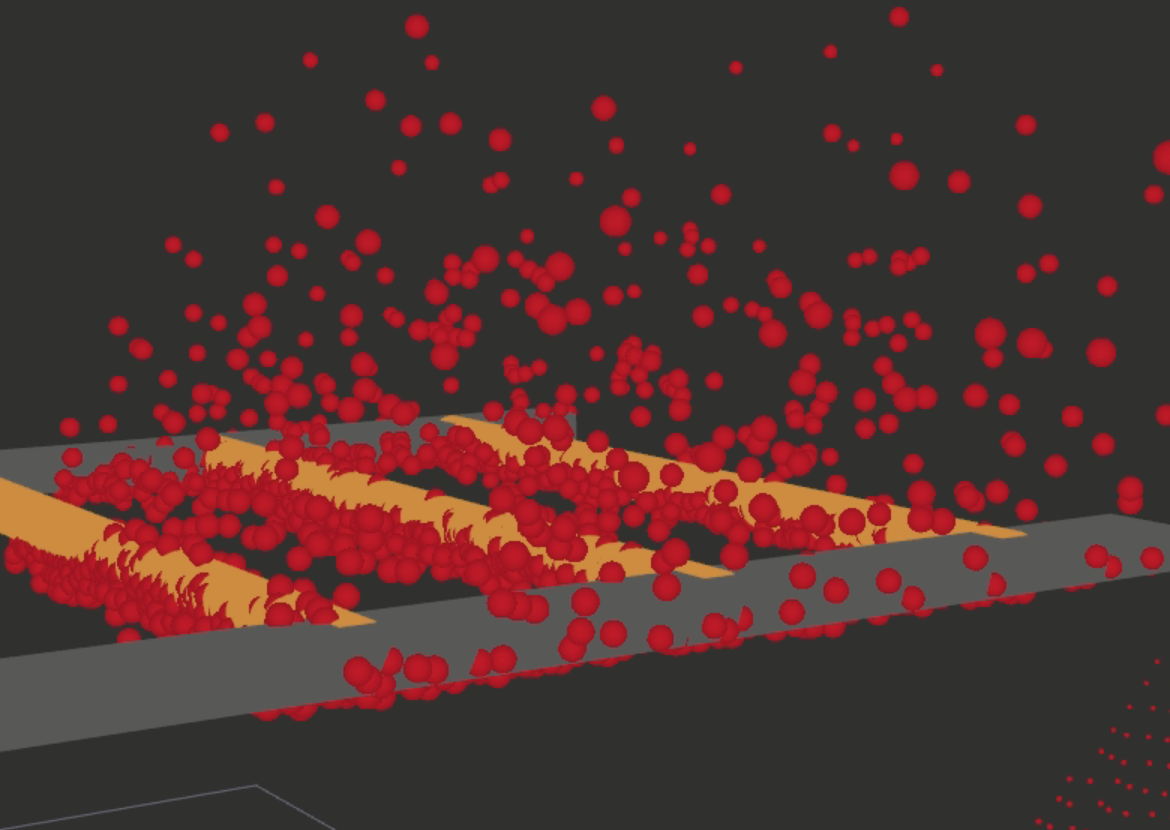}
    \caption{MuJoCo}
    \label{fig:lidar_mujoco}
  \end{subfigure}
  \begin{subfigure}[t]{0.24\textwidth}
    \centering
    \includegraphics[width=\textwidth]{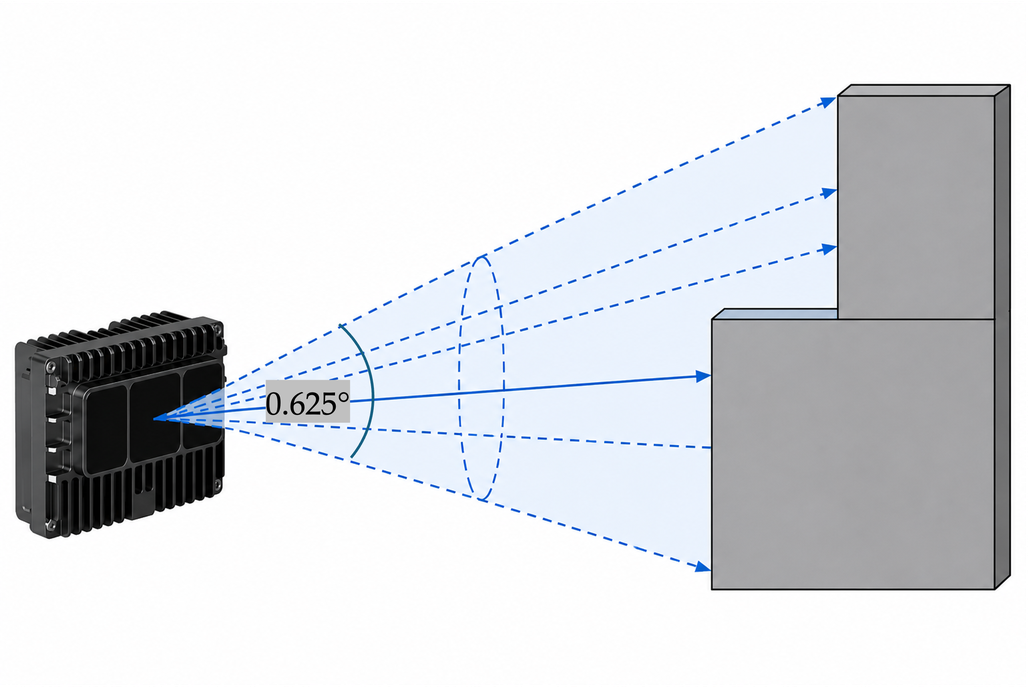}
    \caption{Lidar ray cone model}
    \label{fig:lidar_noise_drawing}
  \end{subfigure}
  \caption{Lidar returns on hardware (a) and in both simulators (b), (c);
    the simulated models reproduce the ray-divergence edge-bleed artifacts
    observed on hardware. (d) The ray-cone model behind both simulators.}
  \label{fig:lidar_artifacts}
\end{figure*}
\subsection{Battery and Thermal Models}
\label{sec:battery_thermal}
\textbf{Battery voltage model.} All joints share one battery whose terminal
voltage sags under load. The summed torque magnitude
$S_t = \sum_j |\tau_{j,t}|$ drives a first-order lag (time constant
$T_\mathrm{rec}$, step $\Delta t$):
\begin{equation}
V_{t+1} = V_t + \tfrac{\Delta t}{T_\mathrm{rec}}\!\left(
  \operatorname{clip}\!\left(V_\mathrm{nom} - k_\mathrm{sag} S_t,\,
  V_\mathrm{min},\, V_\mathrm{nom}\right) - V_t \right).
\end{equation}
The factor $\nu_t = V_t/V_\mathrm{nom}$ rescales each motor's stall torque and
no-load speed, $\tau^\mathrm{sat}_t = \nu_{t-1}\tau^\mathrm{sat}$,
$\dot q^{\max}_t = \nu_{t-1}\dot q^{\max}$, before effort clipping, as
in~\cite{mujoco_dcmotor}. The parameters, derived from the actuator torque
constants and the battery's internal resistance, are tuned so that the
simulated voltage reproduces logs from real jumping sequences. Without this
model, the jump-up drew enough current to brown out the robot; with the
voltage-sag penalty in the reward, no brownout occurred in any hardware
trial.

\textbf{Actuator temperature model.} During brachiation, shoulders and elbows are held near their torque limit for several swings, causing internal temperature increase.
Each of these joints therefore carries a leaky
integrator $I_{j,t}$, a winding-temperature proxy similar
to~\cite{mueller2026olaf} with parameters estimated
from real-world failure cases, which charges with the load ratio
$\rho_{j,t} = |\tau_{j,t}|/\tau_\mathrm{stall}$ and leaks toward zero:
\begin{equation}
I_{j,t+1} = \operatorname{clip}\!\left( I_{j,t} + \Delta t\!\left(
  \tfrac{\rho_{j,t}}{T_\mathrm{ch}} - \tfrac{I_{j,t}}{T_\mathrm{leak}}
\right),\, 0,\, 1 \right).
\end{equation}
A sustained load settles at $I_\infty = \rho\, T_\mathrm{leak}/T_\mathrm{ch}$,
so any sustained load ratio $\rho \geq T_\mathrm{ch}/T_\mathrm{leak} = 0.2$
saturates the integrator. The state is a privileged observation and is
penalized in the reward. Parameters of both models are listed in
Table~\ref{tab:model_params}.

\begin{table}[!t]
\centering
\caption{Battery and actuator-temperature model parameters. The thermal
  model applies to the 18 low-torque actuators (hip yaw, ankles, waist yaw,
  shoulders, elbows, head yaw); both models step at the simulation rate.}
\label{tab:model_params}
\footnotesize
\setlength{\tabcolsep}{3pt}
\begin{tabular}{llr}
  \toprule
  Parameter & Symbol & Value \\
  \midrule
  \multicolumn{3}{l}{\textit{Battery Voltage}} \\
  Nominal voltage & $V_\mathrm{nom}$ & \SI{51}{\volt} \\
  Minimum voltage & $V_\mathrm{min}$ & \SI{30}{\volt} \\
  Sag per unit torque & $k_\mathrm{sag}$ & \SI{0.0375}{\volt\per\newton\per\metre} \\
  Recovery time constant & $T_\mathrm{rec}$ & \SI{0.1}{\second} \\
  \multicolumn{3}{l}{\textit{Actuator temperature}} \\
  Stall torque reference & $\tau_\mathrm{stall}$ & \SI{40}{\newton\metre} \\
  Charge time constant & $T_\mathrm{ch}$ & \SI{1}{\second} \\
  Leak time constant & $T_\mathrm{leak}$ & \SI{5}{\second} \\
  \bottomrule
\end{tabular}
\vspace{-5mm}
\end{table}

\begin{figure*}[!t]
  \centering
  \includegraphics[width=\textwidth]{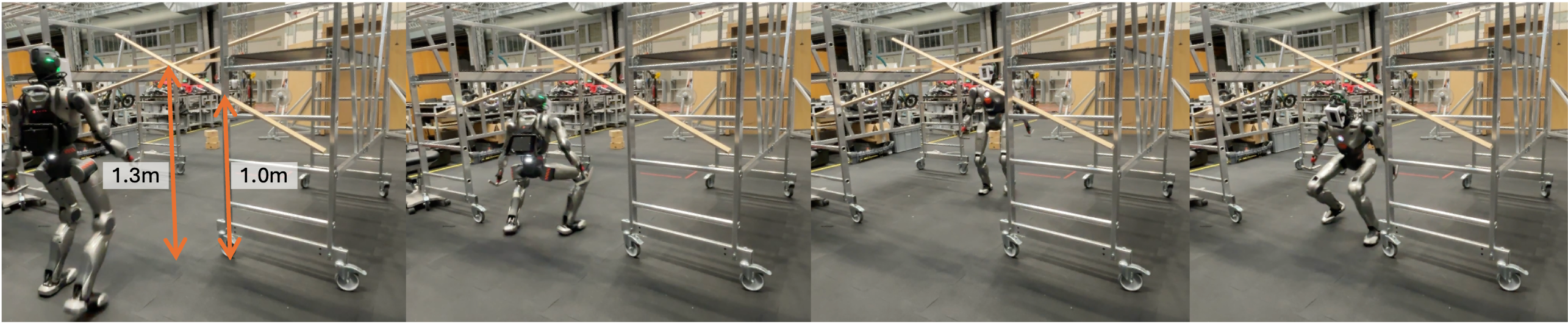}
  \caption{The ducking policy passing under a $2\times2$\,cm wooden slat
    at \SI{1.2}{\metre} clearance.}
  \label{fig:ducking}
\end{figure*}

\subsection{Lidar Noise Model}
\label{sec:lidar_model}
We adopt two noise models for the E1R: a lightweight model optimized for
training throughput, and a higher-fidelity model for sim-to-sim validation.
Their parameters are identified from the sensor itself rather than generic.
The dominant distortion identified on hardware is the divergence of the lidar
ray-cone (Fig.~\ref{fig:lidar_noise_drawing}), which produces edge-bleed
artifacts at depth discontinuities (Fig.~\ref{fig:lidar_real}), the
mixed-pixel effect known from time-of-flight
cameras~\cite{sarbolandi2015kinect}.

The training-time model is implemented in IsaacLab as a custom raycaster written
in Warp~\cite{Macklin_Warp_A_High-performance_2022}. For each pixel, the ray direction is dithered within the
$0.625^\circ$ beam cone, Gaussian range noise ($\sigma = \SI{2}{\centi\metre}$)
is applied to the measured distance, and
outliers are injected through edge detection and background mixing,
mirroring the missing returns the real sensor produces at depth
discontinuities (Fig.~\ref{fig:lidar_isaaclab}). Concretely, a ray is an edge pixel if
its range differs from any 4-neighbor on the 2D grid by more than $0.1$\,m
(no-return neighbors count as far). Each edge pixel is dropped to the invalid
sentinel with probability $0.05$, or mixed with probability $0.20$:
$r' = (1-t)\,r + t\,r_{\mathrm{bg}}$, $t \sim \mathcal{U}(0, 0.7)$, where
$r_{\mathrm{bg}}$ is the farthest 4-neighbor range clamped to the maximum
sensor range; the hit point is re-projected along the ray at $r'$. The model
further randomizes the mounting calibration ($\pm5^\circ$ rotation,
$\pm\SI{2}{\centi\metre}$ position), drops or corrupts $1\%$ of returns,
freezes whole frames with probability $0.1$ for modeling sensor delay, and gates ranges to
$[0.3, 1.5]$\,m, with invalid returns masked as~$-1$.

\subsection{Sim-to-Sim Validation Model}
\label{sec:mujoco_model}
The validation-time model, used in MuJoCo for sim-to-sim evaluation, explicitly
approximates beam divergence: each pixel is rendered by casting multiple rays
stratified inside the beam-divergence cone, modeling the finite footprint of a
solid-state ToF pixel. The per-sample hits $r_s$ are fused with inverse-square
weighting, so nearer surfaces within the cone dominate the return:
\begin{equation}
r \;=\; \frac{\sum_s w_s\, r_s}{\sum_s w_s},
\qquad w_s = r_s^{-2}.
\end{equation}
Pixels at depth discontinuities are dropped to no-return with a probability that
scales linearly with the range gap to a neighbor, saturating at the peak
probability once the gap exceeds the threshold. Spurious returns replace a valid
range with a uniform draw over the sensor's range gate. Each pixel casts
$16$ rays inside the same $0.625^\circ$ cone at \SI{10}{\hertz}, and edge
dropout peaks at $0.5$ for range gaps above \SI{0.3}{\metre}. The resulting
point clouds reproduce the
ray-divergence edge-bleed artifacts observed on hardware
(Fig.~\ref{fig:lidar_mujoco}).

\section{Experiments}
Our experiments address three questions: which perception encoder and
auxiliary supervision best support distillation
(Sec.~\ref{sec:exp_ablation}), whether the learned policies transfer to a
physics engine and sensor model unseen during training
(Sec.~\ref{sec:exp_sim2sim}), and whether they transfer to hardware, for
both the full traversal sequence and ducking under thin overhead obstacles
(Sec.~\ref{sec:exp_hw}). All hardware
experiments run the policy onboard the modified PM-01 of
Fig.~\ref{fig:hero}.

\begin{table}[!t]
  \centering
  \caption{Perception-encoder and auxiliary-loss ablation on the combined
    jump$\to$brachiation$\to$jump-down distillation task. 
    The centerline loss (CL, Sec.~\ref{sec:centerline_probe}) probes the encoders for comparison: values come from decoders fit post hoc to each policy's hidden states. The exception is the first row, which optimizes CL during training as proposed.
    All values $\times10^{-2}$.}
  \label{tab:encoder-ablation}
  \small
  \setlength{\tabcolsep}{3.5pt}
  \begin{tabular}{l r cccc c}
    \toprule
    & & \multicolumn{4}{c}{BC loss $\downarrow$} & \\
    \cmidrule(lr){3-6}
    Encoder & Params & total & jump & brach. & down & CL $\downarrow$ \\
    \midrule
    AME-2 (PC) $+$ aux & $13.8$k & $\mathbf{2.35}$ & $\mathbf{2.48}$
      & $\mathbf{1.95}$ & $\mathbf{3.20}$ & \textcolor{gray}{$0.71$} \\
    AME-2 (PC) & $13.8$k  & $2.43$ & $2.56$ & $2.00$ & $3.26$ & $1.62$ \\
    CNN       & $106.7$k & $2.62$ & $2.73$ & $2.16$ & $3.57$ & $1.79$ \\
    MLP       & $1.31$M  & $2.76$ & $2.92$ & $2.28$ & $3.62$ & $1.85$ \\
    Blind     & --       & $2.90$ & $3.18$ & $2.30$ & $3.51$ & $2.07$ \\
    \bottomrule
  \end{tabular}
\end{table}

\begin{table}[!t]
  \centering
  \caption{Evaluation of the combined perceptive student on the full
    brachiation sequence. $h$: rung height, $s$: bar spacing; number
    of trials in parentheses.}
  \label{tab:combined_eval}
  \small
  \setlength{\tabcolsep}{3.5pt}
  \begin{tabular}{l cc cccc}
    \toprule
    & & & \multicolumn{4}{c}{Success rate [\%] $\uparrow$} \\
    \cmidrule(lr){4-7}
    Configuration & $h$ [m] & $s$ [m]
      & Up & Brach. & Down & Full \\
    \midrule
    \multicolumn{7}{l}{\textit{Sim-to-sim, MuJoCo (10 trials each)}} \\
    & 1.65 & 0.35 & 80 & 100 & 100 & 80 \\
    & 1.70 & 0.35 & 90 & 100 & 100 & 90 \\
    & 1.75 & 0.35 & 90 & 100 & 100 & 90 \\
    & 1.80 & 0.35 & 70 & 100 & 100 & 70 \\
    & 1.85 & 0.35 & 70 & 100 & 100 & 70 \\
    & 1.90 & 0.35 & 70 & 100 & 100 & 70 \\
    \midrule
    \multicolumn{7}{l}{\textit{Real world}} \\
    Ladder A (9) & 1.69 & 0.26 & 100 & 100 & 100 & 100 \\
    Ladder B (2) & 1.72 & 0.31 & 100 & 100 & 100 & 100 \\
    Ladder C (4) & 1.75 & 0.33 & 100 & 75  & 100 & 75 \\
    \bottomrule
  \end{tabular}
  \vspace{-5mm}
\end{table}

\subsection{Encoder and Auxiliary-Loss Ablation}
\label{sec:exp_ablation}
Table~\ref{tab:encoder-ablation} compares point-cloud encoders on the
combined distillation task by their BC loss after the initial distillation stage.
The attention-based AME-2 encoder reaches the lowest loss in every phase
with roughly an order of magnitude fewer parameters than the CNN and two
orders fewer than the MLP; the blind student is worst, confirming that the
task cannot be solved from proprioception alone. Adding the centerline
auxiliary loss (Sec. \ref{sec:centerline_probe}) lowers the BC loss further in every phase, indicating that
explicit geometric supervision aids the recurrent perception backbone.

\subsection{Sim-to-Sim Validation}
\label{sec:exp_sim2sim}
Before hardware deployment, both policies are evaluated in MuJoCo under the
beam-divergence lidar model of Sec.~\ref{sec:mujoco_model}.
Table~\ref{tab:combined_eval} sweeps the rung height for the combined
policy: success remains between $70\%$ and $90\%$ across the
$1.65$--$1.90$\,m range, degrading moderately as the first bar approaches
the limit of the jump. We sweep bar cross-section and underside clearance for the ducking policy: across cylindrical bars with diameter ranging from \SI{0.01}{\metre} to \SI{0.05}{\metre} and clearances from \SI{1.1}{\metre} to \SI{1.5}{\metre}, the policy detects and clears the bar at every clearance the robot fits under, with a $ 100\%$ success rate.

\subsection{Real-World Results}
\label{sec:exp_hw}
We evaluate the full jump-up$\to$brachiation$\to$jump-down sequence on three
ladders (Table \ref{tab:combined_eval}). The robot starts from rest beneath the ladder
and executes the sequence autonomously after the operator specifies a single
goal position in the odometry frame. Onboard pose estimation is provided by
SE(3) lidar-inertial odometry~\cite{se3_lio}, which was the most robust
estimator we tested under task impacts.

Across 15 hardware trials, including runs on weakly supported structures that
shift under load, the policy succeeds in 14, yielding a 93\% full-sequence
success rate. The single failure occurred when the hook did not advance to the
next bar after a successful jump-up. Brachiation speed reaches
\SI{0.5}{\metre\per\second}, comparable to reported human
brachiation speeds~\cite{ani13091438}. Representative motions are shown in
Fig.~\ref{fig:motion}.

For ducking, we test on \SI{2}{\centi\metre}$\times$\SI{2}{\centi\metre}
wooden bars mounted in random orientations unseen during training. The robot
approaches from a standing pose, passes beneath the obstacle without contact,
and recovers to a stable stance. Fig.~\ref{fig:ducking} shows a successful
crossing beneath a \SI{2}{\centi\metre} wooden slat with
\SI{1.2}{\metre} minimum clearance. In 10 passes with randomly placed bars policy successfully cleared the obstacles.

\section{Conclusions and Future Work}
We presented a map-free perceptive control framework that enables a humanoid to
traverse sparse 3D structures directly from solid-state lidar. The key results
are that raw lidar perception can support explosive, contact-accurate
whole-body motions, and that explicit modeling of sensor and actuator effects is
sufficient to transfer these behaviors reliably to hardware while operating near the platform's actuation limits. The same
perception backbone also transfers to ducking under thin overhanging obstacles,
suggesting that the perception setup is useful beyond brachiation.

The current system remains limited to a small set of separately trained task policies, and its robustness to substantially more
diverse geometry remains to be demonstrated.
Extending direct-lidar perception to partially observable structures that require long-horizon spatial memory is an important direction for future work.

\ifanonymous\else
\section*{Acknowledgment}
The authors thank Theo Boldt for their support with hardware experiments, EngineAI for technical support. This work is funded by ETH AI Center.
\fi

\bibliographystyle{IEEEtran}
\bibliography{main}

\end{document}